\documentclass[11pt,a4paper]{article}
\usepackage{times,latexsym}
\usepackage{url}
\usepackage[T1]{fontenc}

\usepackage{graphicx} 
\usepackage{multirow}

\usepackage{caption}
\usepackage{subcaption}

\usepackage{amsmath}
\usepackage{makecell}

\usepackage{hyperref}

\usepackage{comment}

\usepackage{enumitem}

\usepackage[acceptedWithA]{tacl2021v1}

\usepackage{xspace,mfirstuc,tabulary}

\usepackage{booktabs}

\newif\iftaclinstructions
\taclinstructionsfalse 
\iftaclinstructions
\renewcommand{\confidential}{}
\renewcommand{\anonsubtext}{(No author info supplied here, for consistency with
TACL-submission anonymization requirements)}
\newcommand{\instr}
\fi

\iftaclpubformat 

\else

\fi

\title{Benchmarking Linguistic Diversity of Large Language Models}

\author{
  Yanzhu Guo\Thanks{This work was partially done during the author's affiliation with École Polytechnique.} 
  \\
  ALMAnaCH
  \\
  Inria Paris
  \And
  Guokan Shang
  \\
  IFM Paris
  \\
  MBZUAI\\
  \texttt{yanzhu.guo@inria.fr}\\
  \texttt{guokan.shang@mbzuai.ac.ae}\\
  \texttt{chloe.clavel@inria.fr}
  \And
  Chloé Clavel
  \\
  ALMAnaCH
  \\
  Inria Paris
}

\date{}

\begin{document}

\pagestyle{plain}
\thispagestyle{plain}

\maketitle
\begin{abstract}
The development and evaluation of Large Language Models (LLMs) has primarily focused on their task-solving capabilities, with recent models even surpassing human performance in some areas. However, this focus often neglects whether machine-generated language matches the human level of diversity, in terms of vocabulary choice, syntactic construction, and expression of meaning, raising questions about whether the fundamentals of language generation have been fully addressed.
This paper emphasizes the importance of examining the preservation of human linguistic richness by language models, given the concerning surge in online content produced or aided by LLMs.
We adapt a comprehensive framework for evaluating LLMs from various linguistic diversity perspectives including lexical, syntactic, and semantic dimensions. Using this framework, we benchmark several state-of-the-art LLMs across all diversity dimensions, and conduct an in-depth analysis for syntactic diversity. Finally, we analyze how the design, development and deployment choices of LLMs impact the linguistic diversity of their outputs, focusing on the creative task of story generation.

\end{abstract}

\section{Introduction}


Recent Large Language Models (LLMs) have exhibited outstanding capabilities in generating both natural and formal language \citep{brown2020language, touvron2023llama}, while also achieving human-level performance in language understanding, commonsense reasoning, and various other tasks \citep{hendrycks2020measuring}. This has led to evaluations that predominantly focus on these specific abilities \citep{wang2024mmlu}. Meanwhile, other evaluation studies address well-recognized issues in LLMs, such as factuality \citep{maynez-etal-2020-faithfulness}, safety \citep{zhang-etal-2024-safetybench}, and fairness \citep{gallegos-etal-2024-bias}, which remain focal points of ongoing research.
However, there is a notable lack of attention paid to linguistic perspectives, particularly in diversity \citep{guo-etal-2024-curious}, despite the fundamental objective of natural language generation being to produce outputs that are not only \textit{accurate} but also \textit{diverse} \citep{tevet-berant-2021-evaluating}.

Recent studies have highlighted concerns regarding the linguistic diversity of LLM outputs. By comparing human and model-generated content, researchers have shown that models frequently struggle to reflect the nuances and variations found in human expression \citep{shaib2024standardizing,giulianelli-etal-2023-comes}. Additionally, these concerns are reinforced by findings that training language models on synthetic text can lead to a further decline in linguistic diversity \citep{guo-etal-2024-curious}.

In fact, LLMs tend to be inherently conservative in producing diverse content. During training, models undergo homogenization to the most frequent patterns in the training data, where creative outlier narratives, views, styles, and knowledge are often underrepresented \citep{kandpal2023large}. Unlike models, human language production involves a complex interplay of factors that go beyond merely optimizing probabilities \citep{Holtzman2020The}. It is therefore crucial to emphasize evaluating output diversity in language models and systematically consider these metrics to guide future model design, development and deployment decisions.

Currently, principled and comprehensive studies on evaluating linguistic diversity are lacking in the literature \citep{shaib2024standardizing}.
While some works on Natural Language Generation (NLG) report diversity metrics, they typically focus on a single diversity aspect (e.g., lexical diversity~\citep{chakrabarty-etal-2022-help}), often experimenting within a single domain and task (e.g., news summarization ~\citep{shaib2024standardizing}). This narrow focus is problematic since diversity varies across aspects and depends on the domain \citep{guo-etal-2024-curious}.	
Although some efforts have been made to assess the influence of reinforcement learning from human feedback (RLHF) on diversity \citep{kirk2024understanding}, the impact of other key design and development stages—such as model scale, quantization, decoding strategy, and prompt formulation—remains unexplored. 
Additionally, there is a limited understanding of how LLMs develop the capability to generate diverse language through successive pretraining checkpoints.
Ultimately, no study has benchmarked the diversity performance of state-of-the-art LLMs across different aspects and domains.

In this work, we first establish a framework for evaluating linguistic diversity of LLM outputs on a corpus level. We then benchmark six prominent LLMs on five NLG tasks, and compare the diversity of their outputs across three different aspects: lexical, syntactic, and semantic. We place particular emphasis on story generation, the most creative task where linguistic diversity plays a crucial role, conducting a deeper analysis in this context. Specifically, we examine syntactic diversity through a case study comparing the distribution of dependency trees in human-written and LLM-generated texts. Finally, we also investigate how LLM output diversity changes across different development stages, and with varying decisions of deployment.

The main research questions we address are as follows:	
\begin{enumerate}[noitemsep,topsep=0pt,parsep=0pt,partopsep=0pt,leftmargin=*]
    \item What are the key aspects of LLM output diversity, and how can they be evaluated? (See \textsection~\ref{sec:diversity})
    \item How do state-of-the-art LLMs perform in terms of diversity across different tasks? (See \textsection~\ref{sec:benchmarking_results})
    \item How does diversity change during each LLM development stage (e.g., pretraining, supervised fine-tuning (SFT), preference tuning)? (See \textsection~\ref{sec:training_stages})
    \item How do different design (e.g., model scale, training data) and deployment (e.g., decoding strategy, prompt formulation, quantization) choices affect diversity?	(See \textsection~\ref{sec:decoding}, \textsection~\ref{sec:prompt_sensitivity} and \textsection~\ref{sec:scale_quant})
\end{enumerate}

\noindent It is worth noting that we study linguistic diversity in a monolingual context, focusing on the English language. However, the evaluation methodology is language agnostic and could easily be extended to other languages, given that employed NLP toolkits (e.g., dependency parsers, sentence embeddings) exist for the language. Furthermore, our approach to analyzing the influence of various factors on LLM outputs is adaptable to other dimensions, such as linguistic naturalness \citep{guo2024large}.

The code for our research is available at \url{https://github.com/YanzhuGuo/llm-diversity}.

\section{Related Work}
	
In this section, we review methods for evaluating and analyzing linguistic diversity. We define \textit{linguistic diversity} as the natural variation in human language across core linguistic properties, including vocabulary usage, grammatical structures, and semantic nuances. In contrast, a separate line of research focuses on \textit{socio-linguistic diversity} \citep{hayati2023far, lahoti-etal-2023-improving}, which falls beyond the scope of our study.

\subsection{Evaluation of Human Language} \label{subsec:human}
Early metrics for linguistic diversity, proposed by linguists, were developed for studies of language acquisition and language disorder detection. For example, \citet{fergadiotis2013measuring} employed lexical diversity metrics to identify symptoms of aphasia, while \citet{mcnamara2010linguistic} showed that both syntactic complexity and lexical diversity can predict essay quality.		
Another study by \citet{a89efe5d-217a-3260-b2b1-1437ae204234} manually annotated a small corpus of texts produced by second language learners for syntactic features such as syntactic length and clause types, considering their variation as a diversity index. 
However, these metrics are limited to evaluating human-written texts and either focus exclusively on lexical diversity or lack scalability due to the need for manual annotation.	

The evaluation of linguistic diversity in model-generated language has emerged as a relatively recent focus of research. This development is driven, in part, by growing concerns over the increasing online prevalence of model-generated or model-influenced content~\citep{geng2024chatgpt}, prompting questions about whether LLMs can reflect the linguistic richness characteristic of human language~\citep{guo-etal-2024-curious}. However, assessing linguistic diversity is meaningful \textit{only when the generated text meets basic standards of quality}. For instance, a randomly initialized model might produce token sequences with high lexical diversity, but such outputs lack any practical value~\citep{uchendu2023does}. Recent advances in language generation quality have brought model outputs closer than ever to human-like coherence and plausibility, making the evaluation of linguistic diversity more relevant and necessary than before.

\smallskip

\subsection{Evaluation of Generated Language}
To the best of our knowledge, \citet{tevet-berant-2021-evaluating} were the first authors to systematically evaluate diversity in NLG.	
They proposed to create diversity metrics from any two-sentence similarity measure, defining diversity as the inverse of the mean similarity score across all unordered pairs.		
N-gram-based metrics were used to assess form diversity, while model-based metrics like Sentence-BERT similarity measured content diversity.		
They concluded that a notable disparity exists between automatic metrics and human judgment, and that human evaluation of diversity becomes challenging in sets with more than ten sentences.		

Since then, additional metrics have been proposed to capture linguistic diversity, including semantic diversity metrics based on natural language inference \citep{stasaski-hearst-2022-semantic} or semantic entropy \citep{han-etal-2022-measuring}, and syntactic diversity metrics derived from n-grams of Part-of-Speech (POS) tags \citep{giulianelli-etal-2023-comes} or graph similarity kernels of syntax trees \citep{guo-etal-2024-curious}.

Another relevant research direction involves divergence-based metrics that compare the distributions of human-written and machine-generated text. Examples included MAUVE \citep{NEURIPS2021_260c2432}, which leveraged distributions of GPT-2 embeddings, as well as later approaches based on specific linguistic features \citep{guo2024large}. While such metrics do not explicitly measure linguistic diversity, they can offer insights into distributional differences, of which diversity is a key component.

\subsection{Impact of LLMs on Linguistic Diversity}
Diverging from the above research focused on developing methods to evaluate linguistic diversity, another line of work explores how LLM-generated content impact future models or human writing patterns, often demonstrating a decline in diversity. \citet{guo-etal-2024-curious} showed that iteratively training LLMs on synthetic data generated by earlier models leads to a consistent decline in lexical, syntactic, and semantic diversity, especially for tasks requiring high creativity. Similarly, \citet{padmakumar2024does} reported a statistically significant reduction in linguistic diversity when humans write with InstructGPT. This reduction in linguistic diversity is also observed in other contexts: \citet{liang2024mapping} identified a significant frequency shift toward LLM-preferred words in academic writing, while \citet{luo-etal-2024-diverge} reported reduced morphosyntactic diversity in machine translations compared to human translations.

Closely related to our work, \citet{kirk2024understanding} examined how SFT and preference tuning affect LLM generalization and diversity. They found that preference tuning substantially reduces lexical and semantic diversity compared to SFT. Our research also explores the factors that influence diversity while broadening the analysis to include a wider range of diversity aspects, models, tasks and factors. However, our findings on the impact of preference tuning differ from those of \citet{kirk2024understanding}, likely due to differences in task domain, accentuating the importance of contextualizing conclusions.

\section{Metrics for Linguistic Diversity}\label{sec:diversity}

In this section, we present the three types of diversity central to our study: lexical, syntactic, and semantic diversity.

According to \citet{tevet-berant-2021-evaluating}, diversity can be divided into two primary dimensions: form diversity and content diversity. Lexical and syntactic diversity are sub-aspects within form diversity, whereas semantic diversity pertains to content diversity. While additional sub-aspects of form diversity, such as style diversity, exist and could potentially be measured through style representations \citep{soto2024fewshot}, these aspects are generally less interpretable and often overlap with other dimensions of diversity. For instance, style diversity inherently intersects with lexical and syntactic diversity, as stylistic choices typically involve preferences in vocabulary and grammar. Therefore, in this study, we concentrate on the three diversity aspects (lexical, syntactic, and semantic) that are clearly defined, straightforward to interpret, and exhibit relatively low mutual correlation (further detailed in Section~\ref{sec:correlation}).

In terms of evaluation protocol, \citet{kirk2024understanding} distinguish between \textit{across-input diversity} and \textit{per-input diversity}. Across-input diversity refers to the diversity of outputs across different inputs, with only one output generated per input. In contrast, per-input diversity evaluates the capability of the model to produce diverse outputs for a single input.

In our study, we choose to measure across-input diversity, as we focus on general linguistic patterns across a broad range of generations. Formally, given a set of generated outputs $S = \{s_1, s_2, \dots, s_n\}$, we compute $Div(S)$ differently depending on the aspect of diversity: for lexical diversity, $S$ is treated as a set of n-grams, while for syntactic and semantic diversity, $S$ is considered as a set of sentences.

We build on the linguistic diversity evaluation framework and preprocessing methods of \citet{guo-etal-2024-curious}, but shift the focus from studying the effects of recursive synthetic training on OPT \citep{zhang2022opt} to comparing linguistic diversity across a range of state-of-the-art LLMs. We also investigate how various design choices such as model scale and training data, and deployment factors such as decoding strategy, prompt formulation and quantization, impact diversity. In principle, the same research protocol could be extended to examine per-input diversity, allowing for the investigation of uncertainty and variability in text generation \citep{giulianelli-etal-2023-comes}. Although this lies beyond the scope of the current study, it represents a promising direction for future work.

In the following sections, we describe each aspect of diversity and the specific metrics used to assess them.

\begin{table*}[!th]
\centering
\resizebox{\textwidth}{!}{%
\begin{tabular}{@{}llll@{}}
\toprule
                                                   & \textbf{Instruction}              & \textbf{Input}                      & \textbf{Output}                        \\ \midrule
\textbf{Language Modeling (\textbf{LM})}           & Not applicable (no instruction)                              & Block of 128 tokens from Wikipedia  & Prediction of the next block           \\
\textbf{Machine Translation (\textbf{MT})}         & Translate from French to English & News story sentence in French       & Corresponding English translation      \\
\textbf{Summarization (\textbf{Summ})} &
  Summarize the following article &
  Full news article from the BBC &
  Summary of the news article \\
\textbf{Next Utterance Generation (\textbf{NUG})}  & Continue the following dialogue  & Scripted dialogue on general topics & Next utterance of the dialogue         \\
\textbf{Automatic Story Generation (\textbf{ASG})} & Continue the following story     & Story prompt shared by Reddit users & Story continuation based on the prompt \\ \bottomrule
\end{tabular}%
}
\caption{Summary of instructions, inputs, and outputs for benchmarked NLG tasks. Task instructions are placed in the system input when supported, otherwise prepended to the user input.}
\label{tab:tasks}
\end{table*}

\subsection{Lexical Diversity}

Lexical diversity is a measure of the variety of vocabulary used within a text or set of texts. In essence, it assesses the richness or variability of word choices. High lexical diversity indicates a broad range of unique words, while low lexical diversity suggests repetitive or limited vocabulary.

We employ Unique-$n$ \citep{johnson1944studies, templin1957certain}, established for evaluating lexical diversity. It is calculated as the ratio of unique $n$-grams to the total number of $n$-grams. When $n=1$, it is equivalent to Type-Token Ratio \citep{johnson1944studies,templin1957certain}.
We report the average Unique-$n$ across unigrams, bigrams, and trigrams. Originally used in child language research, Unique-$n$ is useful for assessing language development, where a lower value might indicate limited lexical variety \citep{miller1981assessing}. We use the global Unique-$n$ measure rather than the moving average Unique-$n$ because we are interested in the overall diversity capabilities of LLMs across different inputs rather than their performance on individual inputs. Moving average methods might miss global lexical repetitions due to their localized nature \citep{bestgen2023measuring}. To mitigate the influence of output length on Unique-$n$, we always randomly choose 40K samples to constitute the set of $n$-grams for each $n$.

\subsubsection{Syntactic Diversity}
\label{sec:syn}
Syntactic diversity refers to the range and variety of sentence structures used in a text or set of texts. It assesses how flexibly and creatively different grammatical structures, such as phrases, clauses, and sentence types, are employed. High syntactic diversity suggests varied sentence forms, while low syntactic diversity indicates repetitive or simplistic sentence structures.
Syntactic diversity is a crucial but often neglected aspect of language. Exposure to a variety of syntactic structures helps language learners and models develop a richer understanding of language \citep{aggarwal-etal-2022-towards}. Diverse syntactic forms enhance expressiveness and subtlety in text, impacting its style and tone \citep{edwards1998diversity}. While research on syntactic diversity exists, it typically relies on manual annotation, which can be both costly and error-prone \citep{a89efe5d-217a-3260-b2b1-1437ae204234}.

To address this limitation, we employ a graph-based metric for quantifying syntactic diversity \citep{guo-etal-2024-curious}. This metric relies on a neural parser \citep{qi-etal-2020-stanza} to generate dependency trees from sentences, following the universal dependencies framework. In these trees, nodes represent words and edges capture syntactic dependencies, with nodes labeled by the corresponding part-of-speech (PoS) tags. The Weisfeiler-Lehman (WL) graph kernel \citep{10.5555/1953048.2078187, JMLR:v21:18-370} is applied to map these trees into a reproducing kernel Hilbert space, where structurally similar graphs are positioned closer together based on the WL isomorphism test. Syntactic diversity is then computed as the average pairwise distance between these graphs, formalized as: $\text{Div}_{\text{syn}}(S) = \frac{1}{\binom{n}{2}} \sum_{1 \leq i < j \leq n}WL(s_i, s_j)$.

\subsubsection{Semantic Diversity}
Semantic diversity refers to the range and variety of meanings or ideas conveyed within a text or set of texts. It evaluates how broadly and uniquely different concepts, topics, or ideas are expressed, reflecting the depth and scope of the content. Low semantic diversity often indicates repetition or a narrow focus, whereas high semantic diversity typically suggests coverage of a wide array of topics. However, texts should meet a basic quality standard before being evaluated for semantic diversity, since high semantic diversity can also arise from noisy or irrelevant content. Recent studies \citep{tevet-berant-2021-evaluating, stasaski-hearst-2022-semantic} have pointed out that traditional lexical metrics may not fully capture semantic diversity. Similar words can convey different meanings, and different words can convey similar meanings \citep{pmlr-v80-yarats18a}. 

To address this, we first convert sentences into semantically meaningful embeddings using Sentence-BERT \cite{reimers-gurevych-2019-sentence}. Semantic diversity is then quantified as the dispersion of these embeddings in the semantic space, measured by the average pairwise cosine distance (scaled to the range $[0,1]$) between all embedding vectors: $\text{Div}_{\text{sem}}(S) = \frac{1}{\binom{n}{2}} \sum_{1 \leq i < j \leq n}  \frac{d_{\text{cos}}(e(s_i), e(s_j))}{2}$, where $e$ represents Sentence-BERT embeddings.

\begin{figure*}[ht]
    \centering
    \includegraphics[width=0.9\textwidth]{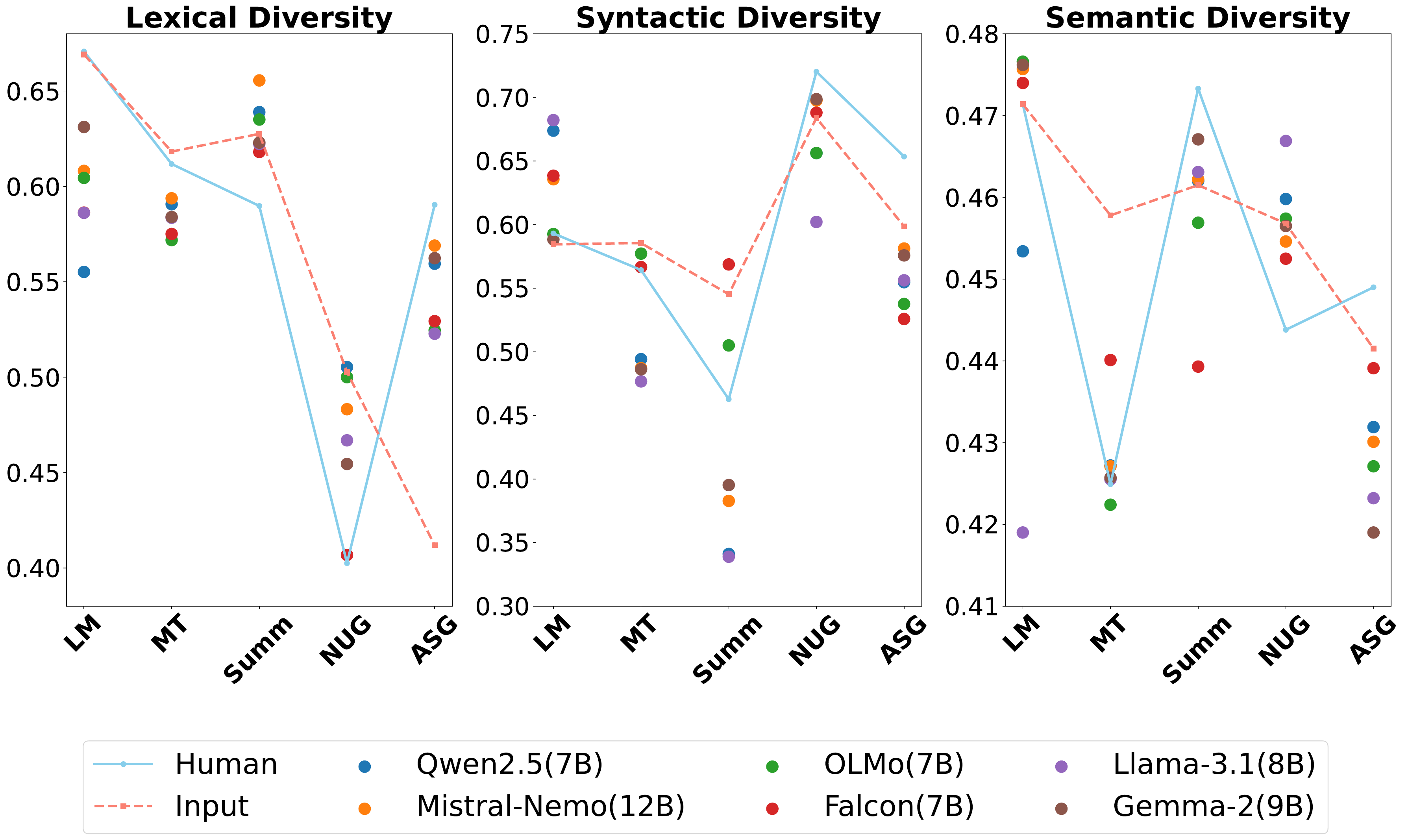}
    \caption{Linguistic diversity benchmarking results for NLG tasks detailed in Table \ref{tab:tasks}.}
    \label{fig:benchmark}
\end{figure*}

\section{Settings for Diversity Benchmarking}
We outline the tasks and models used to establish our linguistic diversity benchmark. We decode the outputs for all tasks and models using a combination of nucleus sampling (t=0.6) and top-k sampling (k=0.9). We further analyze in Section \ref{sec:decoding} the impact of different decoding parameters on output diversity.

\subsection{Generation Tasks}

To effectively compare the linguistic diversity of LLM outputs across various scenarios, we choose five tasks with progressively increasing levels of ``creativity''. The inputs, outputs, and instructions for each task are summarized in Table~\ref{tab:tasks}. To maintain general model behavior and avoid overly influencing responses through prompt design, we keep the instructions minimal.

For each task, we randomly select 10K samples from the original dataset. While our experiments are conducted using a single dataset per task, we deliberately select the most representative for each. Nonetheless, the conclusions drawn from our experiments should be interpreted within the context of these datasets. We now provide a detailed introduction to each task and its associated dataset.

\smallskip
\noindent\textbf{Language Modeling (LM)} involves predicting the next token in a sequence based on the preceding tokens and is fundamental to all NLG applications. We use the Wikitext-2 dataset \citep{merity2017pointer} to evaluate general purpose language modeling. Derived from Wikipedia articles, Wikitext-2 offers a rich corpus with around 2 million tokens across diverse topics. We chunk texts into blocks of 128 tokens and ask the models to predict the next 128 tokens. Language modeling serves as the basis of all other tasks, so it is considered as the least creative.

\smallskip
\noindent\textbf{Machine translation (MT)} aims to transfer text from one language to another while maintaining the original meaning. We use the WMT-14 dataset \citep{bojar-etal-2014-findings} which contains parallel corpora for multiple language pairs. For our experiments, we focus on a subset of this benchmark that includes French-to-English sentence pairs from multiple sources. We classify this task as having a low level of creativity, as the output is expected to convey exactly the same meaning as the input.

\smallskip
\noindent\textbf{Summarization (Summ)} is the process of generating concise summaries of lengthy texts, preserving key information while minimizing redundancy. We use the XLSUM dataset \citep{hasan-etal-2021-xl}, which features news articles in various languages along with their summaries. Our experiments focus on the English portion of the dataset. While we also categorize this task as low in creativity, it allows slightly more flexibility than machine translation, as the model must decide which information to prioritize and include in the summary.

\smallskip
\noindent\textbf{Next Utterance Generation (NUG)} aims to produce natural utterances in conversations while maintaining contextual relevance. For this task, we use the DailyDialog dataset \citep{sai-etal-2020-improving}, a human-curated multi-turn dialogue corpus designed to cover a broad range of topics relevant to everyday interactions. In our setup, the model is always prompted to predict the final utterance based on all preceding dialogue turns. We consider next utterance generation to be a creative task, as there is a large space of possible and coherent utterances in response to a certain dialog context. However, the everyday nature and structure of the dataset place some limits on the level of creativity.

\smallskip
\noindent\textbf{Automatic Story Generation (ASG)} centers on producing engaging and coherent narratives from story prompts or initial contexts. We employ the WritingPrompts dataset \citep{fan-etal-2018-hierarchical}, which comprises prompts and corresponding stories contributed by Reddit users. It includes a wide variety of prompts in different formats, encouraging diverse and creative responses. Among our tasks, we consider story generation to be the most creative, as the prompts typically impose minimal constraints on narrative structure and content, allowing for maximal expressive freedom.

\begin{figure*}[ht]
\begin{minipage}{0.55\linewidth}\centering
\includegraphics[height=9.3\baselineskip]{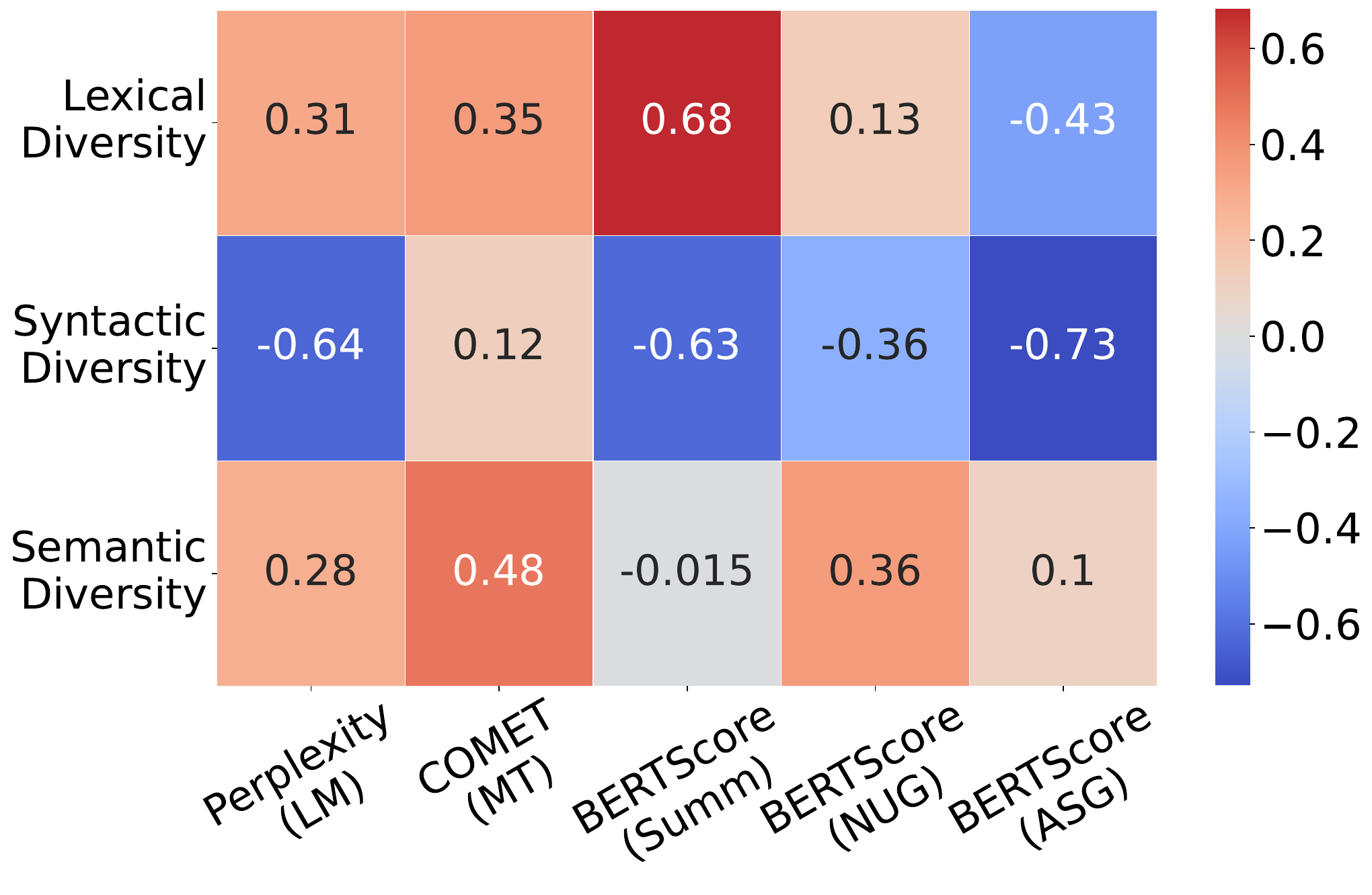}
\caption{Pearson correlation matrix between diversity metrics and quality metrics.}
\label{fig:correlation_qual_div}
\end{minipage}\hfill
\begin{minipage}{0.4\linewidth}\centering
\includegraphics[height=9.3\baselineskip]{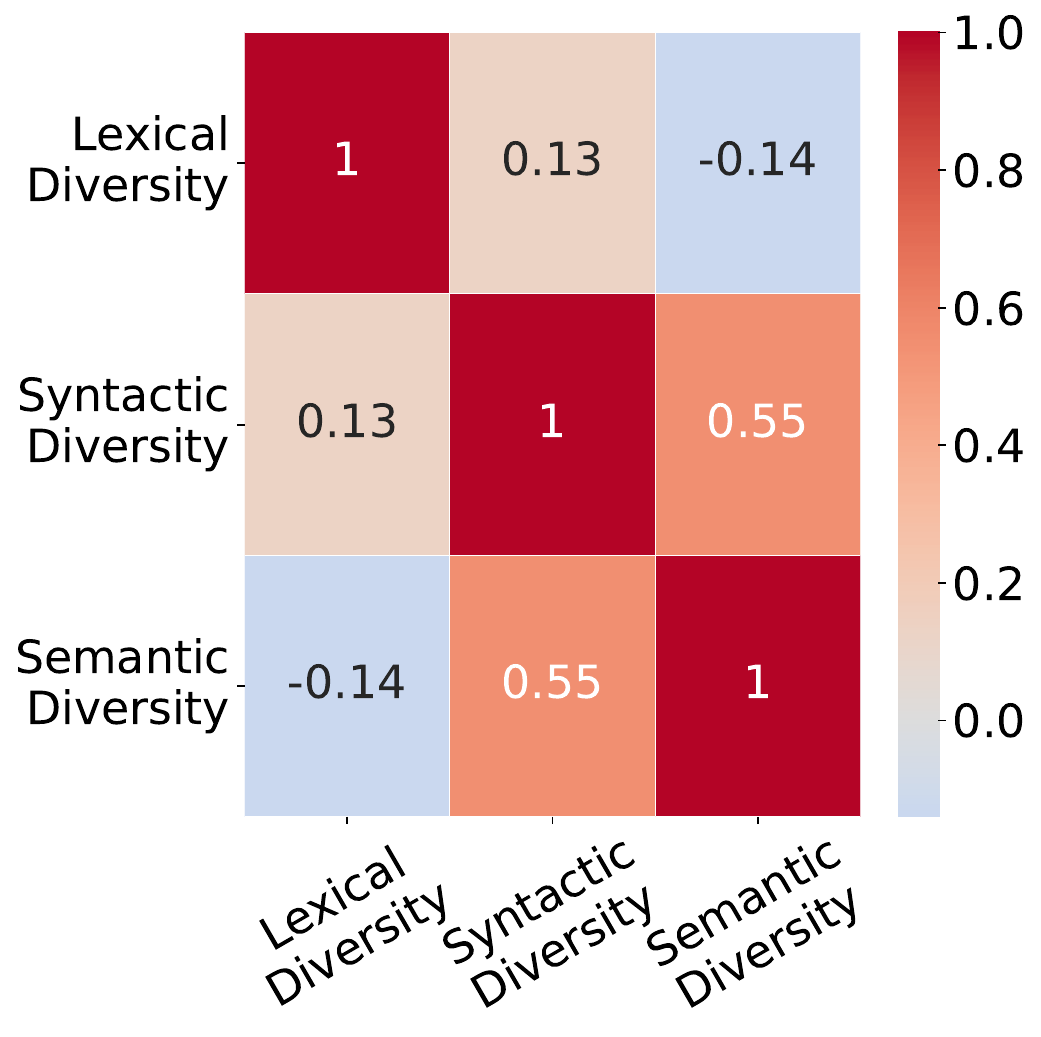}
\caption{Pearson correlation matrix between different diversity metrics.}
\label{fig:correlation_div}
\end{minipage}
\end{figure*}

\subsection{Language Models}
We evaluate the following families of models: Llama \citep{dubey2024llama}, Mistral \citep{jiang2023mistral}, Olmo \citep{groeneveld2024olmo}, Gemma \citep{team2024gemma}, Qwen \citep{yang2024qwen2} and Falcon \citep{almazrouei2023falcon}. The comparison of these models across various key characteristics is provided in Table~\ref{tab:models} in Appendix \ref{app1}.

To ensure comparability, we select the latest version of each model family that is closest in scale to 7 billion parameters. The scale selected for each model is specified in the legend of Figure~\ref{fig:benchmark}. We purposefully include models developed by organizations from different countries to be culturally inclusive. For language modeling, we use base models. For all other tasks, we employ instruction-tuned versions.

\section{Results of Diversity Benchmarking}\label{sec:benchmarking_results}

Figure~\ref{fig:benchmark} visualizes the benchmarking results of linguistic diversity across various tasks. Round dots represent the diversity of model outputs, while solid lines represent human reference outputs. Dashed lines depict the diversity of task-specific inputs (as detailed in Table~\ref{tab:tasks}), reflecting the conditions under which the outputs were generated. Tasks are organized in ascending order of creativity level. The detailed numerical results are provided in Table~\ref{tab:benchmark} in Appendix~\ref{app2}.

For the machine translation task, the inputs are in French; hence, semantic diversity is measured using a multilingual SentenceBERT~\citep{reimers-gurevych-2020-making}, and syntactic diversity is evaluated with a French-specific dependency parser. As a result, these scores may not be directly comparable to those for English. The diversity of human reference outputs serves as a baseline for interpreting whether the model under or over represents the diversity for each task.

In this section, we first analyze metric correlations in Section \ref{sec:CorrelationStudy}, then compare diversity scores across tasks and models in Section \ref{sec:ComparisonTasksModels}. Finally, in Section \ref{sec:ComparisonHumansModels}, we perform a case study on syntactic diversity in story generation, comparing human and model outputs.

\subsection{Correlation Study}\label{sec:CorrelationStudy}

\smallskip
\noindent\textbf{Correlation between diversity and quality}.
We manually verify that all models produce plausible and coherent text that meets the basic requirements for diversity evaluation across all tasks. Building on this, we examine more specific qualities of the model outputs. Figure~\ref{fig:correlation_qual_div} illustrates the correlation between diversity and quality in model outputs, using task-specific automatic metrics as quality indicators. For the language modeling task, perplexity is used to evaluate the model's performance on reference text continuations. For machine translation, we use COMET~\citep{rei-etal-2020-comet}, which takes into account both the source text and reference translation. For the remaining three tasks, BERTScore~\citep{bert-score} is used to measure the relevance between inputs and outputs. Due to the inherently subjective nature of these tasks, automatic metrics generally exhibit weak correlations with human judgments~\citep{liu-etal-2023-g} and should be interpreted cautiously. Nonetheless, we adopt BERTScore as an approximate quality indicator, as embedding-based metrics of this kind demonstrate the strongest system-level correlation with human evaluations among available automatic measures~\citep{chhun-etal-2022-human}.

Our results show a positive correlation between quality and lexical as well as semantic diversity in model outputs. In contrast, syntactic diversity often exhibits negative correlations, where higher syntactic diversity is associated with lower quality scores.
This may be attributed to the tested domains inherently exhibiting low ground-truth syntactic diversity (e.g., in language modeling) or to the limitations of quality metrics in recognizing the value of syntactic variation (e.g., in summarization, automatic story generation, and next utterance generation).
\textit{These findings highlight the need to report diversity metrics alongside quality metrics for comprehensive evaluation, as the relationship between the two is not consistent across tasks or aspects.}

\smallskip
\noindent\textbf{Correlation between diversity aspects}.\label{sec:correlation}
The correlations between different diversity aspects are shown in Figure \ref{fig:correlation_div}, revealing a moderate positive correlation between syntactic and semantic diversity (0.55).
However, lexical diversity shows a weak positive relationship with syntactic diversity (0.13) and a slight negative correlation with semantic diversity (-0.14), indicating that \textit{the richness of vocabulary is independent from the variety of grammatical structures and meaning}.

\begin{table}[!t]
\centering
\resizebox{\columnwidth}{!}{%
\begin{tabular}{@{}lllllll@{}}
\toprule
                   & \textbf{Llama} & \textbf{Mistral} & \textbf{Qwen} & \textbf{Gemma} & \textbf{Falcon} & \textbf{OLMo} \\ \midrule
\textbf{Precision} & 99.20                 & 99.20                     & 99.47               & 99.07               & 99.63              & 99.73            \\
\textbf{Recall}    & 35.20                 & 65.87                     & 75.27               & 37.97               & 75.00              & 39.40            \\ \bottomrule
\end{tabular}%
}
\caption{Comparison of dependency tree distributions between humans and models for the story generation task.}
\label{tab:precision_recall}
\end{table}

\begin{table*}[ht]
\centering
\resizebox{\textwidth}{!}{%
\begin{tabular}{@{}lllll@{}}
\toprule
             & \multicolumn{2}{c}{\textbf{Human}}         & \multicolumn{2}{c}{\textbf{Language Models}}           \\
             & \textbf{POS tag n-gram} & \textbf{Example} & \textbf{POS tag n-gram} & \textbf{Example} \\ \midrule
\textbf{n=3} & (ADV, ADV, ADP)         & right along with & (PRON, NOUN, ADJ)       & her voice soft   \\
\textbf{n=4} & (VERB, ADP, DET, NOUN)          & picking up the pieces         & (NOUN, CCONJ, NOUN, PRON)           & carvings and symbols that          \\
\textbf{n=5} & (DET, NOUN, ADP, DET, NOUN)     & the cackling of the fire      & (PRON, NOUN, VERB, ADP, NOUN)       & its feathers stained with blood    \\
\textbf{n=6} & DET, ADJ, NOUN, ADP, DET, NOUN) & the old woman down the street & (ADJ, NOUN, ADP, NOUN, CCONJ, NOUN) & particular focus on time and space \\ \bottomrule
\end{tabular}%
}
\caption{Examples of syntactic patterns favored by either humans or models are illustrated using n-grams of POS tags. Human patterns are derived from human dependency trees that are not within the model dependency tree neighborhoods, while model patterns have high frequency in model dependency trees and low frequency in human dependency trees.}
\label{tab:examples}
\end{table*}


\subsection{Comparison Across Tasks and Models} \label{sec:ComparisonTasksModels}

We now examine the results in Figure \ref{fig:benchmark} to assess human diversity results across tasks, compare model diversity against human diversity, and finally evaluate the diversity performance across different models.

\smallskip
\noindent\textbf{Human output diversity}.
\textit{Human-level diversity varies across tasks, with no clear correlation observed among different aspects.}
Notably, utterances in human dialogs exhibit the lowest lexical diversity and the highest syntactic diversity, unlike the written text present in the remaining four tasks. The low lexical diversity may be attributed to the conversations being specifically scripted for English learners to practice daily-life dialog. These dialogs focus on generic topics, leading to a limited range of vocabulary. In contrast, the high syntactic diversity can be explained by the inherent spontaneity of conversational language, where different speakers tend to vary significantly in their use of syntactic structures \citep{healey2014divergence, dubuisson-duplessis-etal-2017-automatic}. Human summaries show limited lexical and syntactic variation but exhibit the highest semantic diversity, suggesting a narrow range in form but a broad range in content. In contrast, human translations score lowest in semantic diversity, reflecting the restricted topical scope of the source texts. Wikipedia-based language modeling displays high topic diversity, while human-written stories tend to be diverse across all three dimensions.

\smallskip
\noindent\textbf{Model output diversity}.
\textit{LLMs lack diversity compared to humans for tasks demanding high levels of creativity, such as story generation.}
Overall, the scores of different LLMs across tasks and diversity aspects tend to resemble each other, potentially due to the use of similar development procedures, architectures, and datasets. However, this remains an assumption, as most LLM developers do not fully disclose their training data or protocols, even when the models themselves are open-weight. The extent to which LLMs under- or over-represent diversity compared to humans varies significantly by the task domain. For the task of story generation, which demands the highest levels of creativity and freedom of expression, LLMs consistently lag behind humans in all three diversity aspects. In contrast, for tasks like next utterance generation, LLMs surpass human references in both lexical and semantic diversity. This discrepancy arises because the DailyDialog dataset focuses on generic, everyday topics designed for English language learning, while LLMs, unconstrained by this context, frequently steer conversations toward more complex topics.

\smallskip
\noindent\textbf{LLM comparisons}.
While the overall performance of the models appears to be similar, in-depth comparisons showcase notable differences.		
\textit{Models pretrained on fewer tokens, such as Falcon and OLMo, consistently generate outputs with lower lexical diversity.} Specifically, Falcon and OLMo are pretrained on 1.5T and 2.7T tokens, respectively, compared to Llama-3.1, which is trained on 15T tokens.
However, this effect is not observed for syntactic or semantic diversity.
\textit{Models with less strict data filtration exhibit greater diversity in creative tasks}, such as story generation.	
For example, Qwen2.5, which filters data exclusively for quality, exhibits significantly higher diversity in story generation across all aspects compared to Llama-3.1, Gemma-2, and OLMo, whose data is extensively filtered for quality, privacy, and safety.

\begin{figure*}[ht]
    \centering
    \includegraphics[width=0.85\textwidth]{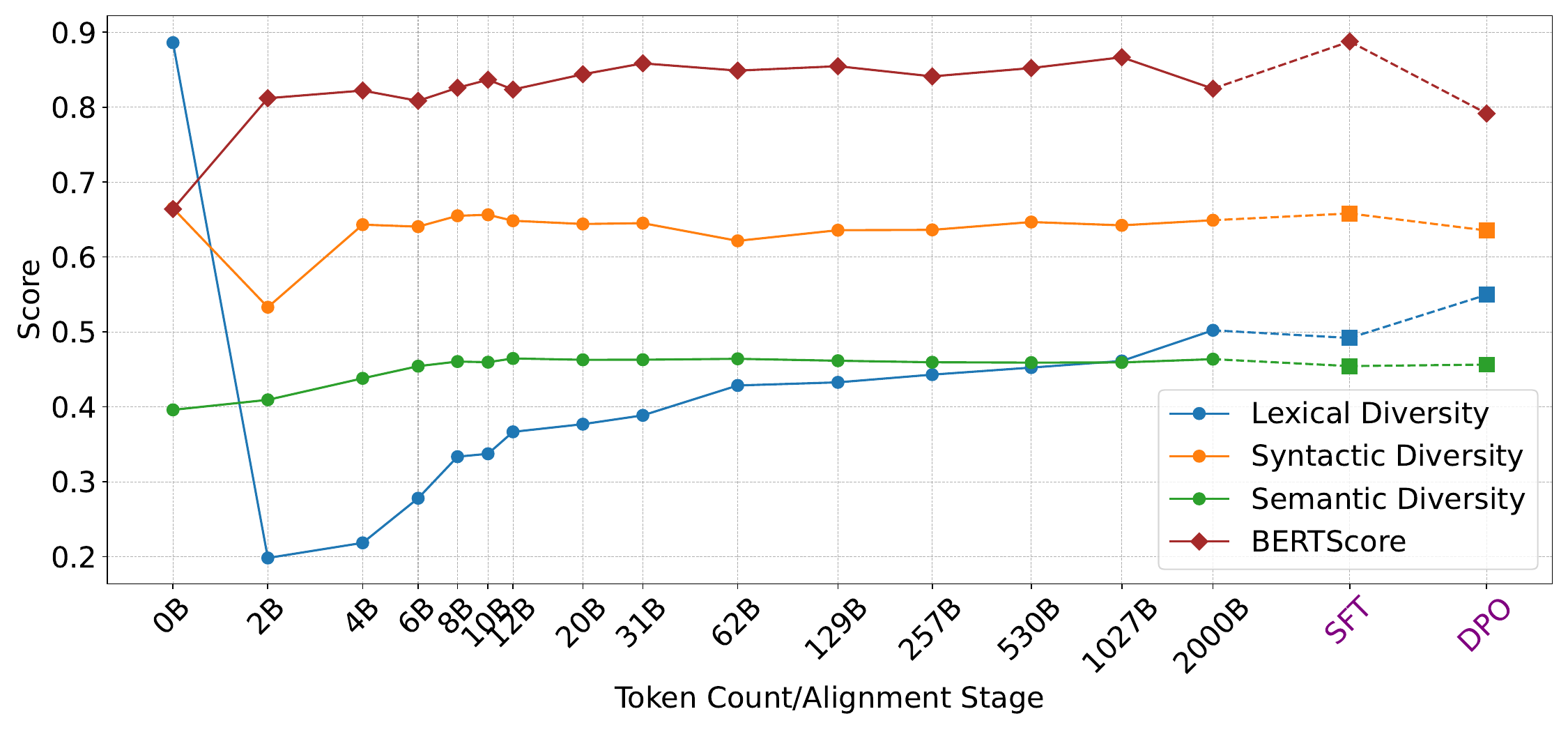}
    \caption{Linguistic diversity metrics after different LLM training stages. The pretraining stage is broken into various steps with increasing token counts, which are presented on a log scale for visualization. Experiments are conducted with the OLMo model on the story generation task.}
    \label{fig:olmo}
\end{figure*}

\begin{figure}[!t]
    \centering
    \includegraphics[width=0.95\columnwidth]{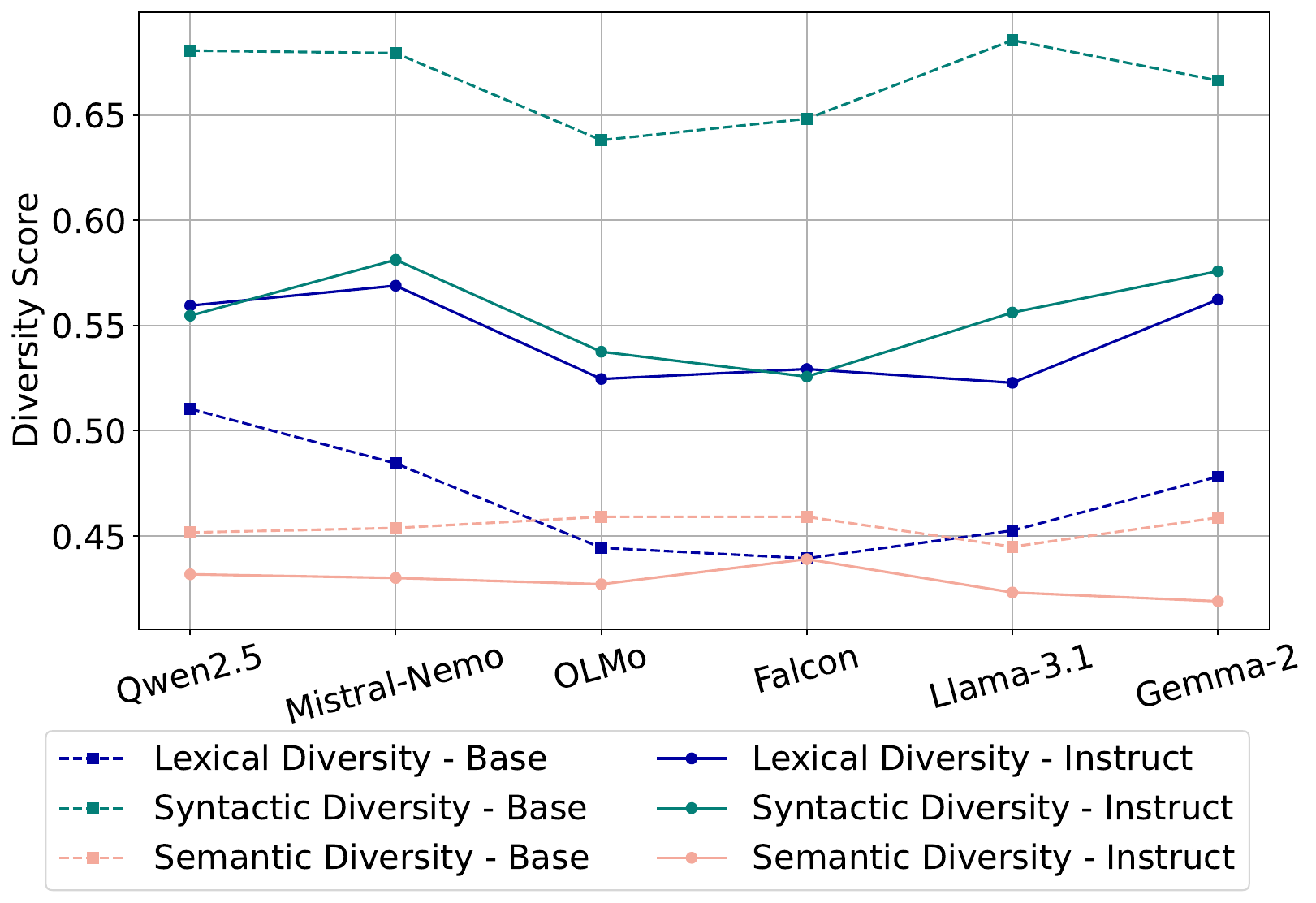}
    \caption{Impact of instruction tuning.}
    \label{fig:base}
\end{figure}

\begin{figure*}[h]
    \centering
    \includegraphics[width=0.85\textwidth]{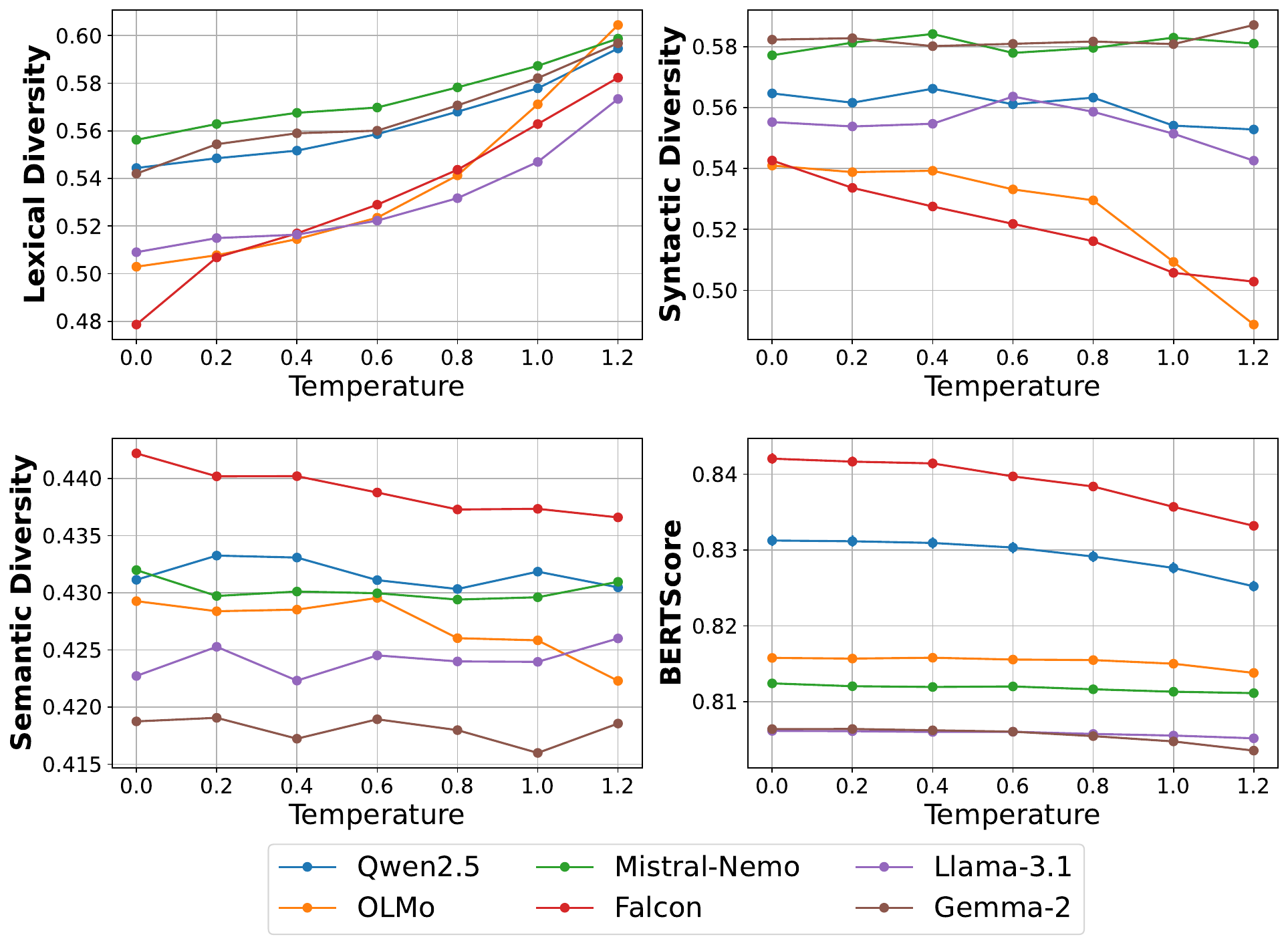}
    \caption{Impact of decoding parameters. Experiments are conducted on the story generation task.}
    \label{fig:decoding}
\end{figure*}

\subsection{Comparing Syntactic Diversity Between Humans and Models}\label{sec:ComparisonHumansModels}

To further compare humans and models, we conduct a case study on syntactic diversity using dependency tree distribution. Syntactic diversity is chosen as it is less explored than lexical and semantic diversity. Moreover, syntactic patterns reflected by POS tag n-grams are more generalizable than lexicon n-grams and more interpretable than semantic embeddings.

We adopt the Precision-Recall framework proposed by \citet{le-bronnec-etal-2024-exploring}. This framework relies on GPT-2 embeddings, followed by Principal Component Analysis (PCA) and K-means clustering, to estimate the supports of text distributions. In our study, we replace the original GPT-2 embeddings with the implicit distribution of dependency tree embeddings induced by the WL graph kernel. Precision is defined as the proportion of dependency trees from model-generated text that lie within the support of dependency trees from human-written text. A high precision indicates that the model-generated structures are more plausible and human-like, thus reflecting their quality. Recall, on the other hand, measures the proportion of dependency trees from human-written text that fall within the support of the model-generated distribution. A high recall suggests that the model captures the full diversity of human-written structures. The method for computing pairwise distances between dependency trees is described in Section~\ref{sec:syn}, which serves as the basis for constructing the distance matrix. All other hyper-parameters remain consistent with the original work \citep{le-bronnec-etal-2024-exploring}.

Table~\ref{tab:precision_recall} presents the precision and recall scores for all evaluated models on the story generation task. The results reveal that all models exhibit near-perfect precision, indicating that almost all generated sentences are syntactically plausible. In contrast, recall scores are substantially lower than precision scores across all models, revealing their limited capacity to capture the full breadth of human syntactic diversity. \textit{This points to a notable gap between models and humans in syntactic diversity for the story generation task where high creativity is required}.

To further illustrate these findings, Table~\ref{tab:examples} lists examples of syntactic patterns (POS tag n-grams) that are frequently found in human dependency trees but are missing from the model-generated ones. Conversely, we also identify syntactic patterns that models over-generate but are less common in human outputs. Recent studies~\citep{shaib-etal-2024-detection} indicate that models often memorize syntactic templates encountered during pretraining, which are rarely overwritten during SFT and preference tuning. This suggests that the observed gap in syntactic patterns may stem from a mismatch between pretraining and downstream task domains. For instance, in the pretraining corpus of OLMo, over 80\% of the data originates from web pages in Common Crawl, while less than 0.3\% comes from Project Gutenberg books, one of the only sources potentially aligned with the narrative style required for story generation \citep{soldaini-etal-2024-dolma}.

\section{Factors Influencing LLM Diversity}

In this section, we explore key factors that may influence the diversity of LLM outputs. The factors under consideration include pretraining token counts, instruction tuning, decoding parameters, prompt formulation, model scale, and quantization. For decoding parameters, prompt formulation and instruction tuning, we conduct experiments across all models. We employ OLMo for assessing the impact of pretraining token counts since it provides full access to its pretraining datasets and model weights at various checkpoints throughout its development. Since OLMo models are available in only two sizes, we additionally leverage Qwen2.5 models \citep{yang2024qwen2} to investigate the effects of model scale and quantization.

All experiments in this section are conducted on the story generation task, where linguistic diversity plays a central role. Its minimal input constraints and strong emphasis on creativity make it an ideal benchmark for evaluating linguistic diversity. Moreover, as shown in Figure~\ref{fig:benchmark}, all models fall significantly short of human performance in terms of diversity on this task, highlighting the importance of identifying which factors contribute to this gap. \textit{We emphasize that the conclusions drawn in this section are specific to the story generation task and should not be generalized to broader LLM behavior without further investigation.}

\subsection{Impact of Training Stages}\label{sec:training_stages}
\smallskip
\noindent\textbf{Pretraining}.
We choose OLMo, pretrained on the Dolma corpus \citep{soldaini-etal-2024-dolma}, to study the evolution of linguistic diversity during pretraining. This is because OLMo is the only model in our benchmark with publicly available intermediate checkpoints during pretraining. The results are presented in Figure~\ref{fig:olmo}. Intitially, lexical diversity is exceptionally high, as expected for an untrained model that generates random tokens. This metric drops sharply after the first checkpoint (2B tokens) but then gradually increases throughout the pretraining process, without reaching saturation. In contrast, syntactic diversity also experiences a sharp decline early on; however, it saturates much more quickly, fluctuating within a narrow range afterward. Semantic diversity shows a steady increase from the beginning but also saturates relatively quickly. \textit{These observations suggest that while increasing training data generally improves lexical diversity, alternative strategies are needed to enhance syntactic and semantic diversity.}

\smallskip
\noindent\textbf{Instruction tuning}.
We now move on to study the impact of instruction tuning on linguistic diversity. After pretraining, OLMo underwent supervised fine-tuning (SFT) on Tulu v2 \citep{ivison2023camelschangingclimateenhancing} and direct preference optimization (DPO) \citep{rafailov2023direct} on Ultrafeedback \citep{cui2024ultrafeedback}, with DPO applied on top of SFT. We observe that SFT has minimal impact on any diversity metric, while DPO leads to a decrease in syntactic diversity and an increase in lexical diversity, potentially reflecting characteristics of the SFT and DPO datasets.

Since all models in our benchmark provide both base and instruction-tuned versions, we extend our analysis to assess the impact of instruction tuning across the full set. As shown in Figure~\ref{fig:base}, the results mirror those observed for OLMo: instruction-tuned versions show higher lexical diversity compared to their base counterparts but exhibit reductions in syntactic and semantic diversity. Notably, the decline in syntactic diversity is more pronounced than that in semantic diversity. 
\textit{These findings indicate that while additional training—regardless of the stage—enhances vocabulary richness, aligning models with human preferences tends to constrain them to a narrower range of grammatical structures and meanings.}

\begin{figure}[!t]
    \centering
    \includegraphics[width=0.95\columnwidth]{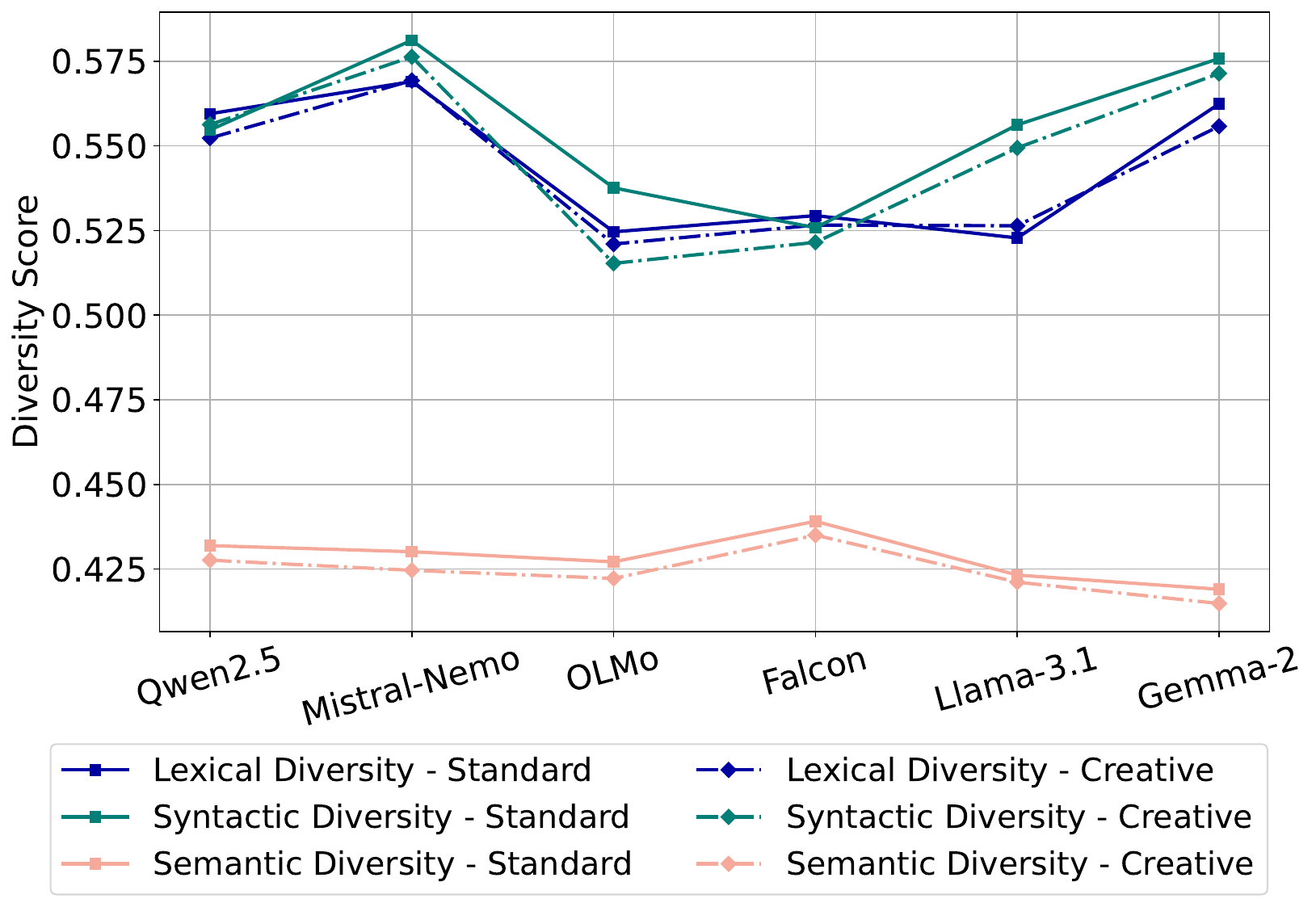}
    \caption{Impact of prompt formulation.}
    \label{fig:prompt_sensitivity}
\end{figure}

\subsection{Impact of Decoding Parameters}\label{sec:decoding}

Achieving a balance between quality and diversity in LLM outputs is a known challenge, as there is often a trade-off between these two aspects \citep{Caccia2020Language}. The choice of decoding strategy plays a crucial role in controlling this trade-off~\citep{zhang-etal-2021-trading}. Here, we investigate how varying the decoding temperature affects the outputs in the story generation task, with results visualized in Figure~\ref{fig:decoding}. Output quality is estimated based on their relevance to the inputs, using BERTScore as a metric.

\textit{The results show that increasing the temperature---making decoding less restrictive---leads to greater lexical diversity, with only a minor reduction in relevance to the inputs.} It might be due to the creative nature of the story generation task that the quality-diversity trade-off is so subtle. For syntactic diversity, while most models show fluctuating performance within a certain range, some exhibit a clear downward trend, specially OLMo and Falcon, which are trained on significantly fewer tokens compared to the other models. However, no consistent trends are observed for semantic diversity metric as decoding parameters change. This aligns with the observations of \citet{tevet-berant-2021-evaluating}, which indicate that adjusting decoding parameters tends to affect the form of the text rather than its meaning.

Furthermore, we note that, across most models, the relative ranking of diversity scores remains stable as the temperature varies. \textit{This suggests that conducting experiments with a fixed temperature is sufficient for consistent evaluation.} Based on our findings, we set the temperature to 0.6 for all other experiments. Figure~\ref{fig:decoding} shows that at a temperature of 0.6, the relevance to the inputs remains relatively high while diversity scores significantly improve compared to lower temperatures.

\begin{figure}[!t]
    \centering
    \includegraphics[width=0.95\columnwidth]{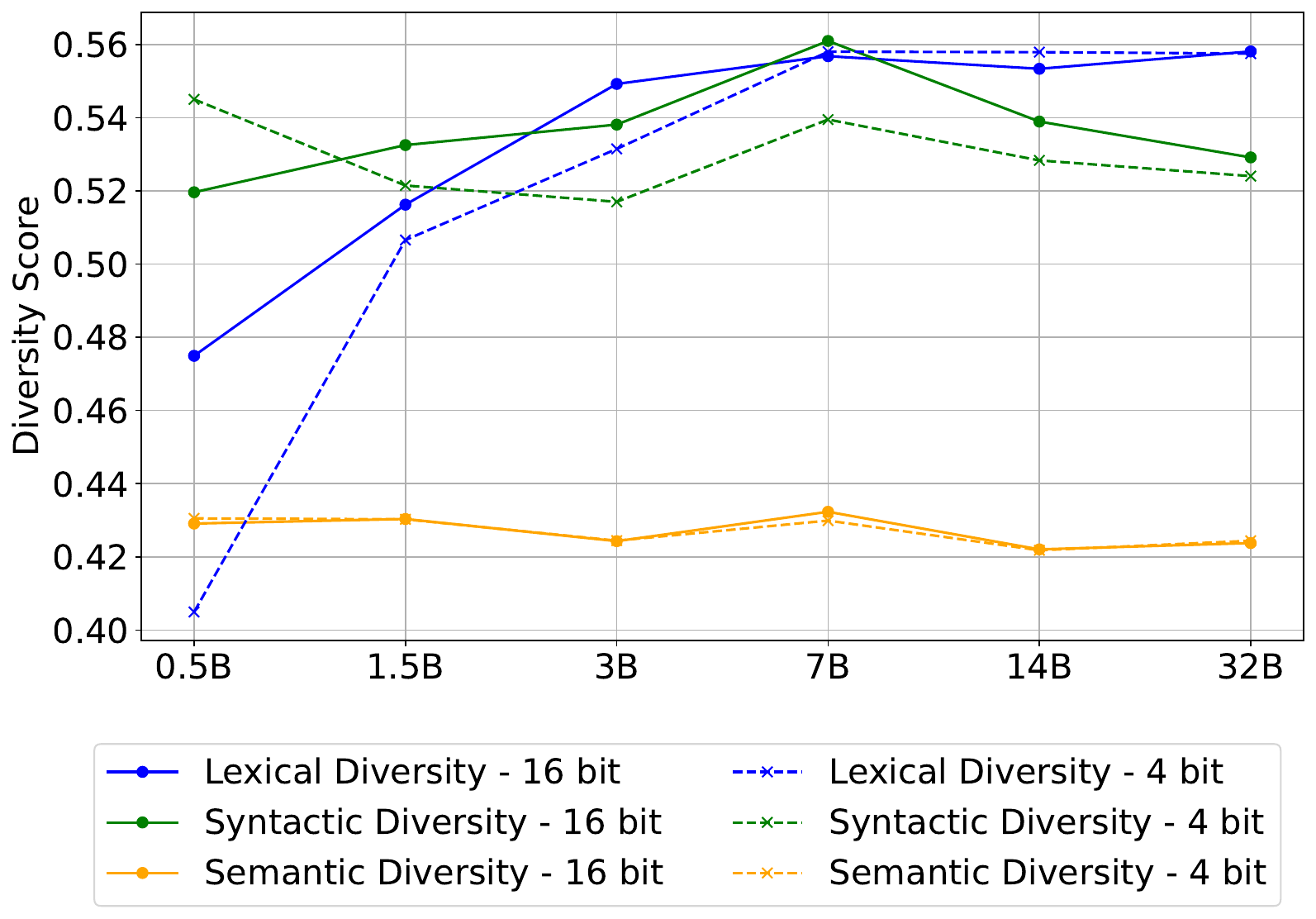}
    \caption{Impact of model scale and quantization.}
    \label{fig:scale_quant}
\end{figure}

\subsection{Impact of Prompt Formulation}\label{sec:prompt_sensitivity}

Previous studies have established that LLMs exhibit considerable sensitivity to prompt formulations, particularly affecting their performance on discriminative downstream tasks \citep{sclar2024quantifying, wahle-etal-2024-paraphrase}. Here, we explore whether the linguistic diversity of stories generated by LLMs is similarly influenced by variations in the formulation of prompts. We conduct experiments across the full range of models, and the results are depicted in Figure~\ref{fig:prompt_sensitivity}.
The solid lines represent results obtained using the standard prompt, consisting of the task-specific instruction ``Please continue the following story'' combined with sample-specific inputs from the WritingPrompts dataset. To evaluate prompt sensitivity, we modify the prompt to explicitly encourage creativity by changing the instruction to ``Please continue the following story and be as creative as possible''. Results from these modified prompts are shown as dash-dot lines in Figure~\ref{fig:prompt_sensitivity}.

\textit{Our analysis indicates that altering the prompt formulation has minimal impact on the diversity of generated stories across all three evaluated aspects.} This suggests that the linguistic patterns exhibited by LLMs across creative generations represent inherent model characteristics that are less sensitive to prompt variations compared to accuracy-based performance on discriminative tasks. Consequently, enhancing linguistic diversity in LLM outputs through straightforward prompt engineering alone would be challenging.

\subsection{Impact of Model Scale and Quantization}\label{sec:scale_quant}

We now study the impact of model scale on linguistic diversity with the Qwen2.5 model. Qwen2.5 has been released in various sizes, ranging from 0.5B to 72B parameters. Due to computational resource constraints, we limit our exploration of linguistic diversity to models up to 32B parameters. The results are presented in Figure \ref{fig:scale_quant}. \textit{We observe that lexical diversity consistently increases with model size, while semantic diversity remains stable throughout}. In contrast, syntactic diversity remains relatively stable overall but exhibits an initial increase followed by a decline, peaking at 7B parameters, \textit{indicating that scaling up is not always the solution to higher linguistic diversity}.

We further investigate the impact of post-training quantization on linguistic diversity. We quantize the Qwen2.5 models of various scales to 4-bit precision with the bitsandbytes library\footnote{\href{https://huggingface.co/docs/bitsandbytes/index}{https://huggingface.co/docs/bitsandbytes/index}},
whereas the original models were run with bf16. As shown in Figure \ref{fig:scale_quant}, \textit{quantization does not affect semantic diversity but reduces both syntactic and lexical diversity}. The reduction in lexical diversity is more pronounced in smaller models, while the effect on syntactic diversity becomes more evident in larger models. \textit{This finding suggests that quantization has greater impact on the diversity of form rather than content.}

\section{Conclusion}
Our study offers crucial insights into the linguistic diversity of current LLMs. By leveraging a comprehensive evaluation framework focused on lexical, syntactic, and semantic diversity, we provide a fresh perspective beyond traditional quality metrics. Our analysis reveals that, despite the impressive capabilities of LLMs in generating coherent and plausible text, there is a significant gap when it comes to replicating the linguistic richness of human language for creative tasks such as story generation. Furthermore, we find that factors like pretraining data volume, instruction tuning, decoding strategies, model scale, and quantization significantly influence diversity metrics. In particular, while instruction tuning improves lexical diversity, it constrains syntactic and semantic diversity, indicating a narrowing of expressive flexibility. These findings raise an important concern: as LLMs become more prevalent in content creation, their outputs may trend towards homogenization, risking a loss of linguistic richness. Our research highlights the necessity of a more holistic and forward-looking approach in developing language models, one that prioritizes the preservation of linguistic diversity alongside optimizing performance metrics.

\section*{Acknowledgments}
We thank Professor Michalis Vazirgiannis for providing the computational resources that supported this project. This research was partially funded by the ANR-23-CE23-0033-01 SINNet project and the ANR-TSIA HELAS chair.

\bibliography{tacl2021, anthology}
\bibliographystyle{acl_natbib}

\onecolumn
\appendix

\section{Comparison of Benchmarked LLMs} \label{app1}

\begin{table*}[!th]
\small
\centering
\setlength{\tabcolsep}{1pt} 
\renewcommand{\arraystretch}{1} 
\scalebox{0.9}{
    \begin{tabular}{c|ccccccc}
    \toprule
    &  \textbf{Llama-3.1-8B}&  \textbf{Mistral-NeMo-12B}&  \textbf{Qwen2.5-7B}&  \textbf{Gemma-2-9b}&  \textbf{Falcon-7b}&  \textbf{OLMo-7B}\\
    \midrule
    \textbf{Organization}& Meta & Mistral & Alibaba & Google & TII & Ai2\\
    \textbf{Country}& USA & France & China & USA & UAE & USA\\
    \textbf{Open weights}& yes & yes & yes & yes & yes & yes\\
    \textbf{Open data}& no & no & no & no & partially & yes\\
    \textbf{Tokenization} & BPE (Tiktoken) & BPE (Tiktoken) & BPE & SentencePiece & BPE & BPE\\
    \textbf{Vocabulary size}& 128K & 128K & 151K & 256K & 65K & 50K\\
    \textbf{\#tokens}& 15T & unknown & 18T & 8T & 1.5T & 2.7T \\
    \textbf{Data filter}& \makecell{quality,\\ privacy, safety} & unknown & quality & \makecell{quality,\\ privacy, safety} & quality & \makecell{quality,\\ privacy, safety}\\
    \textbf{Synthetic data}& post-training & unknown & pre/post-training & post-training & unknown & post-training\\
    \textbf{Multilinguality}& yes & yes & \makecell{yes\\ (over 29 languages)} & not in particular & \makecell{yes\\ (Latin alphabet)}& no\\
    \textbf{Alignment}& \makecell{rejection sampling,\\ SFT, DPO} & SFT & SFT, DPO & SFT, PPO & SFT & SFT, DPO\\
    \textbf{Release date}& July 2024 & July 2024 & September 2024 & June 2024& May 2023 & February 2024\\
    \bottomrule
    \end{tabular}
}
    \caption{Comparison of benchmarked LLMs.}
    \label{tab:models}

\end{table*}

We provide a comprehensive comparison of the LLMs included in our benchmark in Table~\ref{tab:models}, highlighting several key characteristics relevant to their design, development, and deployment.

\section{Detailed Results of Linguistic Diversity Benchmarking} \label{app2}

\begin{table}[!ht]
\centering
\resizebox{\textwidth}{!}{%
\begin{tabular}{@{}l|lllllllll@{}}
\toprule
 &
   &
  {\color[HTML]{3531FF} \textbf{Human}} &
  {\color[HTML]{CB0000} \textbf{Input}} &
  \textbf{Qwen2.5} &
  \textbf{Mistral-Nemo} &
  \textbf{OLMo} &
  \textbf{Falcon} &
  \textbf{Llama-3.1} &
  \textbf{Gemma-2} \\ \midrule
 &
  \textbf{LM} &
  {\color[HTML]{3531FF} 67.08} &
  {\color[HTML]{CB0000} 66.92} &
  55.52 &
  60.82 &
  60.45 &
  58.63 &
  58.62 &
  \textbf{63.12} \\
 &
  \textbf{MT} &
  {\color[HTML]{3531FF} 61.18} &
  {\color[HTML]{CB0000} 61.83} &
  59.07 &
  \textbf{59.38} &
  57.19 &
  57.51 &
  58.36 &
  58.40 \\
 &
  \textbf{Summ} &
  {\color[HTML]{3531FF} 58.98} &
  {\color[HTML]{CB0000} 62.76} &
  63.90 &
  \textbf{65.56} &
  63.51 &
  61.81 &
  62.23 &
  62.30 \\
 &
  \textbf{NUG} &
  {\color[HTML]{3531FF} 40.25} &
  {\color[HTML]{CB0000} 50.27} &
  \textbf{50.53} &
  48.32 &
  50.00 &
  40.68 &
  46.69 &
  45.45 \\
\multirow{-5}{*}{\textbf{\begin{tabular}[c]{@{}l@{}}Lexical\\ Diversity\end{tabular}}} &
  \textbf{ASG} &
  {\color[HTML]{3531FF} 59.04} &
  {\color[HTML]{CB0000} 41.19} &
  55.95 &
  \textbf{56.90} &
  52.46 &
  52.94 &
  52.28 &
  56.24 \\ \midrule
 &
  \textbf{LM} &
  {\color[HTML]{3531FF} 59.31} &
  {\color[HTML]{CB0000} 58.45} &
  67.39 &
  63.57 &
  59.26 &
  63.85 &
  \textbf{68.22} &
  58.83 \\
 &
  \textbf{MT} &
  {\color[HTML]{3531FF} 56.43} &
  {\color[HTML]{CB0000} 43.47} &
  49.42 &
  48.71 &
  \textbf{57.72} &
  56.66 &
  47.67 &
  48.62 \\
 &
  \textbf{Summ} &
  {\color[HTML]{3531FF} 46.27} &
  {\color[HTML]{CB0000} 54.52} &
  34.10 &
  38.27 &
  50.50 &
  \textbf{56.87} &
  33.88 &
  39.52 \\
 &
  \textbf{NUG} &
  {\color[HTML]{3531FF} 72.03} &
  {\color[HTML]{CB0000} 68.39} &
  65.63 &
  69.76 &
  65.63 &
  68.80 &
  60.21 &
  \textbf{69.87} \\
\multirow{-5}{*}{\textbf{\begin{tabular}[c]{@{}l@{}}Syntactic\\ Diversity\end{tabular}}} &
  \textbf{ASG} &
  {\color[HTML]{3531FF} 65.35} &
  {\color[HTML]{CB0000} 59.86} &
  55.47 &
  \textbf{58.12} &
  53.76 &
  52.58 &
  55.62 &
  57.58 \\ \midrule
 &
  \textbf{LM} &
  {\color[HTML]{3531FF} 47.14} &
  {\color[HTML]{CB0000} 47.14} &
  45.34 &
  47.57 &
  \textbf{47.66} &
  47.40 &
  41.90 &
  47.62 \\
 &
  \textbf{MT} &
  {\color[HTML]{3531FF} 42.49} &
  {\color[HTML]{CB0000} 33.58} &
  42.72 &
  42.71 &
  42.24 &
  \textbf{44.01} &
  42.55 &
  42.57 \\
 &
  \textbf{Summ} &
  {\color[HTML]{3531FF} 47.33} &
  {\color[HTML]{CB0000} 46.15} &
  46.20 &
  46.22 &
  45.69 &
  43.93 &
  46.31 &
  \textbf{46.71} \\
 &
  \textbf{NUG} &
  {\color[HTML]{3531FF} 44.38} &
  {\color[HTML]{CB0000} 45.68} &
  45.98 &
  45.46 &
  45.74 &
  45.25 &
  \textbf{46.69} &
  45.65 \\
\multirow{-5}{*}{\textbf{\begin{tabular}[c]{@{}l@{}}Semantic\\ Diversity\end{tabular}}} &
  \textbf{ASG} &
  {\color[HTML]{3531FF} 44.90} &
  {\color[HTML]{CB0000} 44.15} &
  43.19 &
  43.01 &
  42.71 &
  \textbf{43.91} &
  42.32 &
  41.90 \\ \bottomrule
\end{tabular}%
}
\caption{Linguistic diversity benchmarking results for NLG tasks detailed in Table \ref{tab:tasks}. For each type of diversity, the highest model score for each task is highlighted in bold.}
\label{tab:benchmark}
\end{table}

We present the detailed results of our linguistic diversity benchmarking experiments in Table~\ref{tab:benchmark}. These results are also visualized in Figure~\ref{fig:benchmark}.

\end{document}